\documentclass[letterpaper]{article} 
\usepackage{aaai24}  
\usepackage{times}  
\usepackage{helvet}  
\usepackage{courier}  
\usepackage[hyphens]{url}  
\usepackage{graphicx} 
\usepackage{natbib}  
\usepackage{caption} 
\usepackage{algorithm}
\usepackage{algorithmic}
\usepackage{color}

\usepackage{graphicx}
\usepackage{amsmath,amssymb,amsthm,amsfonts}
\usepackage{acronym}
\usepackage{enumitem}
\usepackage{xspace}

\usepackage[switch]{lineno}
\usepackage{url}
\usepackage{multirow}
\usepackage{booktabs}
\usepackage{mathrsfs}
\usepackage{arydshln}

\usepackage{array}
\usepackage{colortbl}
\usepackage{pifont}

\makeatletter
\DeclareRobustCommand\onedot{\futurelet\@let@token\@onedot}
\def\@onedot{\ifx\@let@token.\else.\null\fi\xspace}
\def\eg{\emph{e.g}\onedot}

\def\ie{\emph{i.e}\onedot}

\def\etc{\emph{etc}\onedot}

\makeatother

\acrodef{nlp}[NLP]{natural language processing}
\acrodef{plm}[PLM]{pretrained language model}
\acrodef{sota}[SOTA]{state-of-the-art}
\acrodef{bs}[BS]{Beam Search}
\acrodef{mhs}[MHS]{Metropolis-Hastings Sampling}
\acrodef{hs}[HS]{Hybrid Search}
\acrodef{uas}[UAS]{unlabeled attachment score}
\acrodef{dda}[DDA]{Directed Dependency Accuracy}
\acrodef{sota}[SOTA]{state-of-the-art}
\acrodef{pos}[POS]{part-of-speech}

\def\objs{O}
\def\properties{P}
\def\actions{A}
\def\action{a}
\def\transition{M}
\def\initial{I}
\def\goals{G}
\def\goal{g}
\def\task{t}

\def\tasks{T}
\def\state{s}
\def\states{S}
\def\trajectory{\tau}

\def\model{\texttt{{Helpy}}\xspace}
\def\taskname{\texttt{Autonomous Helping}\xspace}
\def\sr{SR}
\def\exec{Exec}
\def\gcr{GCR}

\usepackage{newfloat}

\usepackage{listings}
\usepackage[misc]{ifsym}

\usepackage{newfloat}
\usepackage{listings}
\DeclareCaptionStyle{ruled}{labelfont=normalfont,labelsep=colon,strut=off} 
\floatstyle{ruled}
\newfloat{listing}{tb}{lst}{}
\floatname{listing}{Listing}

\nocopyright

\title{Get the Ball Rolling: Alerting Autonomous Robots When to Help to Close the Healthcare Loop}
\author{
    Jiaxin Shen\textsuperscript{\rm 1},
    Yanyao Liu\textsuperscript{\rm 1},
    Ziming Wang\textsuperscript{\rm 1}
    Ziyuan Jiao\textsuperscript{\rm 2}
    Yufeng Chen\textsuperscript{\rm 1}
    Wenjuan Han\textsuperscript{\rm 1}\footnote{Corresponding Author}
}
\affiliations{
    \textsuperscript{\rm 1}Beijing Jiaotong University, Beijing, China \\
    \textsuperscript{\rm 2} Beijing Institute for General Artificial Intelligence, Beijing, China \\

}

\usepackage{bibentry}

\begin{document}

\maketitle

\begin{abstract}
To facilitate the advancement of research in healthcare robots without human intervention or commands, we introduce the \taskname Challenge, along with a crowd-sourcing large-scale dataset. The goal is to create healthcare robots that possess the ability to determine when assistance is necessary, generate useful sub-tasks to aid in planning, carry out these plans through a physical robot, and receive feedback from the environment in order to generate new tasks and continue the process. Besides the general challenge in open-ended scenarios, \taskname focuses on three specific challenges: autonomous task generation, the gap between the current scene and static commonsense, and the gap between language instruction and the real world. Additionally, we propose \model, a potential approach to close the healthcare loop in the learning-free setting. 
\end{abstract}

\section{Introduction}
\begin{figure}[!t]
    \flushleft
    \includegraphics[width=1\linewidth]{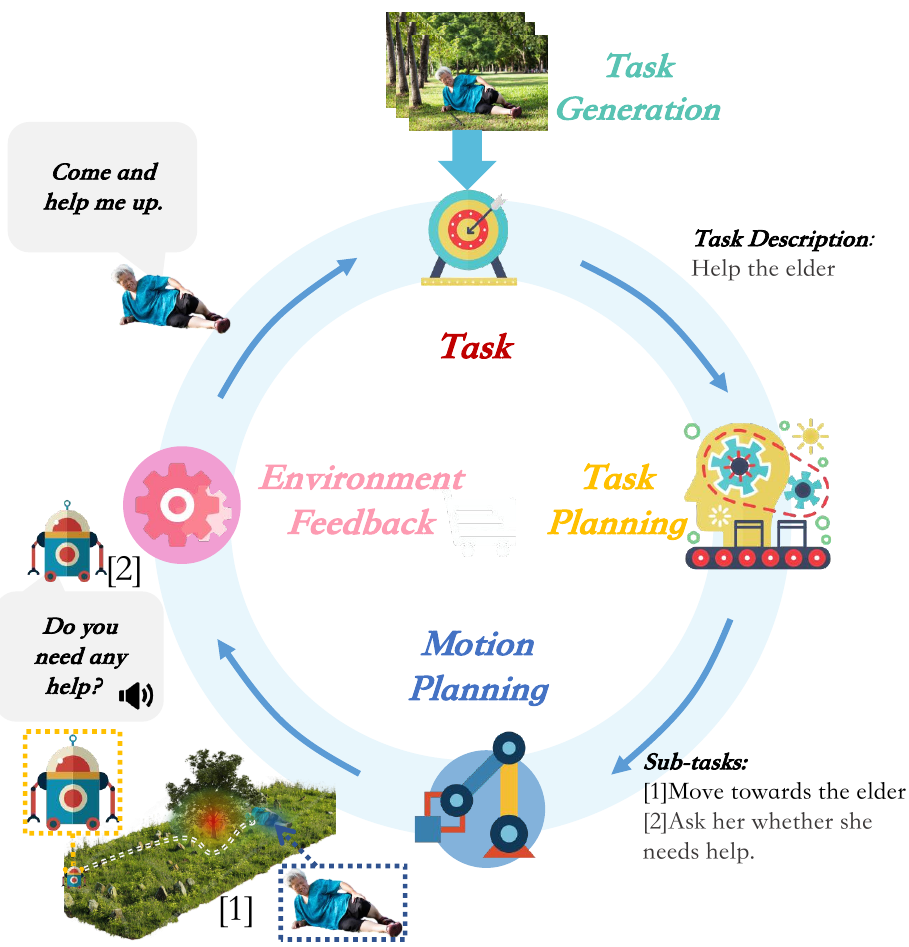}
    \caption{\taskname aims to enable autonomous capabilities in healthcare scenarios. In previous work, explicit instructions (namely, tasks) were given. Here, the \textcolor[RGB]{178,40,28}{\textit{Task}} should be generated to start the whole loop. Besides, both \textcolor[RGB]{253,180,8}{\textit{Task Planning}} and \textcolor[RGB]{54,91,183}{\textit{Motion Planning}} are included.
    }
    \label{fig:motivation}
\end{figure}
A service robot is a type of robot that performs useful tasks for humans or equipment, as defined by the International Organization for Standardization (ISO), excluding industrial automation applications. These robots are gaining popularity due to their ability to provide several advantages over traditional human labor, ultimately freeing up human workers~\citep{schraft2000service,lu2020service,holland2021service}.
One prevalent type of service robot is the healthcare robot, particularly with the growing concern over the global aging population. As the number of elderly individuals living alone continues to rise, so does the demand for healthcare robots for elders and patients~\citep{holland2021service}. 

The key challenge and essential requisition of a healthcare robot, that differs from other robots, is that it should be able to provide services without human intervention or commands~\citep{bekey2005autonomous}. Just as ISO 8373 term states, robots need ``a certain degree of autonomy'', which is the capability to perform expected goals based on the current state and sensing, without human intervention. Take Fig.~\ref{fig:motivation} of a person falling down as an example. Although the person may not say anything directly to give direct commands, the robot must be able to sense its surroundings, make decisions about when to provide help, make plans by generating helpful and supportive sub-tasks (namely, task planning; represented as natural language commands --- ``\textit{Move towards the elder}'' and ``\textit{Ask her whether she needs help.}''), execute these plans (namely, motion planning), and perceive the physical environment feedback, followed by generating new tasks to continue the loop. 
To close the loop between abstract language instructions and robot actions, many studies explore the problem of grounding language instructions for the next-step executing~\cite{tellex2020robots,singh2023progprompt}: leveraging lexical analysis to parse the instructions~\citep{tellex2011understanding,kollar2010toward,bollini2013interpreting} and leveraging large language models (LLMs) to decompose the instructions into an action sequence~\citep{huang2022language,ahn2022can,zeng2022socratic}. More recently, \citet{huang2023voxposer} transforms the given instruction to robot trajectories for execution.
However, existing approaches typically rely on given instruction. The question then arises: How can we ``get the ball rolling'' when robots do not have instructions to finish the following steps of the loop?

In this work, we dive into closing the loop for healthcare robots and introduce a new challenge -- \taskname Task -- aiming at building a robot that should be able to autonomously generate and complete tasks without human intervention. 
Besides the general challenge of autonomous robots in open-ended scenarios, we highlight three main challenges specific to our \taskname: 
\begin{enumerate}[leftmargin=*,noitemsep,nolistsep]
\item \emph{Autonomous task generation}: Alerting robots when assistance is needed is a significant challenge that has received little attention in previous research.
\item \emph{Gap between current scene and static commonsense}: 
\taskname adopts the first-person vision, which is the perspective of human observation and perception of the world~\citep{li2023eye,liu20224d}. Even though fusing current first-person vision information and commonsense knowledge is challenging, a surge of interest is witnessed by the community. 
\item \emph{Gap between language instruction and the real world}: While prior arts~\citep{puig2020watch,singh2023progprompt} are thought to learn generalizable knowledge from text, it remains unclear how to utilize such knowledge to enable embodied robots to act in the physical world.
\end{enumerate}

To address these challenges, we propose a potential approach for \taskname, named as \model. \model leverages the previous success of ProgPrompt~\citep{singh2023progprompt}, which is a close-domain approach and needs instructions from humans. We extend it in an open-ended scenario and fully autonomous setting. Specifically, as sketched in Fig.~\ref{fig:model}, it perceives the current environment in the first-person vision through the vision-language model (VLM) and maintains the perception in the state database. 
Already at this stage, we combine the \emph{current scene and static commonsense} by simultaneously having the real-time video streaming attend to the static foundation models (namely, VLM). 
This makes the robot aware of the current scene, effectively exploiting the physical world without pre-defined objects, addressing the second challenge. 
On top of this combination, we regularly inquire VLM using prompts if there are any situations that require help so as to autonomously generate new tasks when help is needed. We refer to this as {\em autonomous task generation}. Here, we further split the generated task into sub-tasks by using LLM for specifying compositional goals, referred to as task planning. 
To this end, \model performs action planning, which typically is achieved by motion planners instead of pre-defined primitives like \citet{huang2023voxposer}, to bridge the \emph{gap between language instruction and the real world}.


In summary, our contributions are three-fold: (i) A new challenge -- \taskname Challenge for healthcare robots, and corresponding dataset (Sec.~\ref{sec:task}); (ii) A novel method --- \model, to close the healthcare loop (Sec.~\ref{sec:approach}); (iii) Promising performance for learning-free setting (Sec.~\ref{sec:experiments}).


\section{\taskname Challenge}\label{sec:task}
\begin{figure*}[!t]
    \centering
    \includegraphics[width=1.0\linewidth]{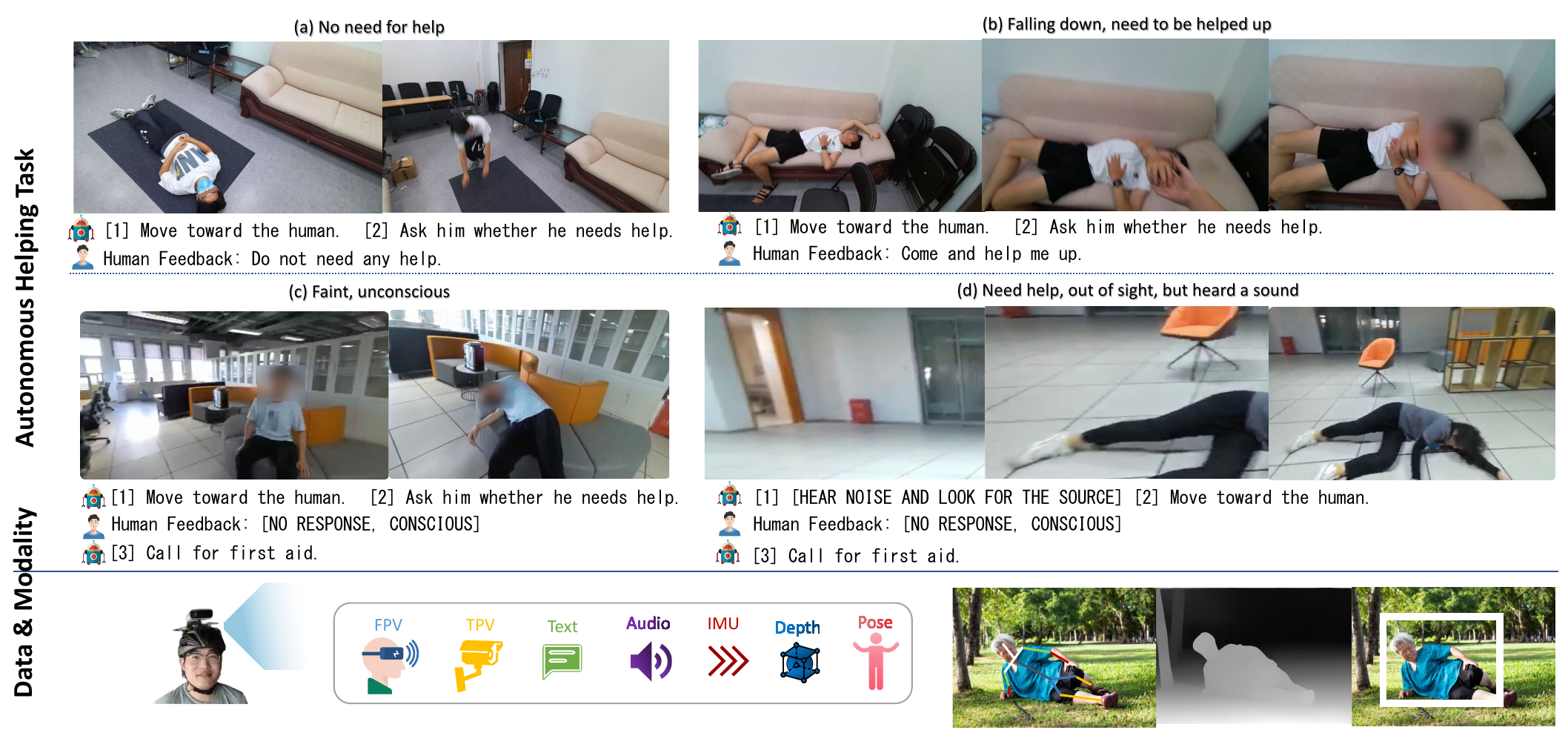}
   \caption{Overview of the \taskname Challenge and the dataset. \taskname includes various scenarios: need help (b) or not (a), conscious (b) or unconscious (c), in (c) or out of sight (d). The robot is inspired to sense the  surroundings, communicate with humans, and act without supervision in an unlearning setting. We collected a multi-modal dataset with seven modalities including FPV (First-person view), TPV (Third-person view), Text (Communication text), Audio, IMU (Inertial measurement unit), Depth (Depth image), as well as varied labels (\eg, poses, segmentation, tasks, functions \etc).}
    \label{fig:task}
\end{figure*}

\taskname Challenge (as sketched in Fig.~\ref{fig:task}) aims at constructing healthcare robots that should be able to sense their surroundings, make decisions about when to provide help, make plans by generating helpful and supportive sub-tasks (namely, task planning), execute these plans (namely, motion planning), and perceive the physical environment feedback, followed by generating new tasks to continue the loop. 
Given the first-person RGBD video streaming along with other sensors, a robot is expected to complete the whole loop without human intervention or instructions. During the whole loop, it must also be able to communicate with humans and modify the plan at any time based on the current state.

\subsection{Problem Setup}
\label{sec:task_formulation}
Given the first-person RGBD video and historical state, the robot predicts if the helping behaviors from the robot are needed at this current moment as a Yes/No question. Once the answer is Yes, it means the robot has the motivation to help and should take action. 
Unlike most previous challenges performing high-level planning in a pre-defined domain~\cite{fikes1971strips,garrett2020pddlstream,jiang2018task}, we adopt the open-ended domain without pre-defined objects or primitives. 
Since the pre-defined domain makes the algorithms that can be hard to scale in environments with many feasible actions and objects~\cite{puig2018virtualhome,shridhar2020alfred} due to the physical world with large branching factors. We use language as the task planning space and the 6-DoF pose as the motion planning space. 

We formulate the task planning as the tuple $\langle \objs, \properties, \actions, \transition, \initial, \goals, \tasks \rangle$. $\objs$ is a set of all the objects in the current scene as well as the video history, $\properties$ is a set of properties of the objects that also informs object affordances. Notable, $\properties$ may be variables for example distances or 6-DoF pose. $\actions$ is a set of executable actions that change depending on the current environment state defined as $\state \in \states$. $\state$ is the current environment state with the assigned $\properties$ and $\states$ is a set of all possible assignments. The transition model is defined as $\transition:~\states~\times~\actions~\rightarrow~\states$. $\initial$ and $\goals$ denote the initial states and goal states, respectively. The robot does not have access to the specific task, but it actively generates a task $\task \in \tasks$ with a high-level task description to achieve the goal state, $\goal \in \goals$. The robot can only do the following five actions: talk to the person, pick up an object, put the object somewhere, and hold or release the human's arm. 


\textcolor{black}{Consider the generated task of helping people to stand up,  \textit{\task= hold\_release(object, target\_place, agent)}. 
Task relevant objects \textit{human\_arm} $\in \objs$ will have properties \textit{object\_pose} modified during action execution.
For example, the initial 6-DoF pose of arm $(rot_{init}, tr_{init}, 1)$, action $\action = hold(\textit{human\_arm}, \textit{target\_pose}, \textit{robot\_hand})$, where $\textit{target\_pose}= (rot_{tgt}, tr_{tgt}, 1)$, will change the state (the 6-DoF pose of the \textit{human\_arm}) from $(rot_{init}, tr_{init}, 1)$  to $(rot_{tgt}, tr_{tgt}, 1)$,if $\action$ is admissible, i.e., $\exists (\action, \state, \state')\; s.t.\; \action \in \actions \wedge \state, \state' \in \states \wedge \transition(\state, \action) = \state'$.
Instead, we focus on sub-tasks where the decomposition $\task \to (\task _1, \task _2, \dots, \task _n)$ is given by a task planner (\eg, an LLM-based planner or a search-based planner).} 


Consider a motion-planning problem given a sub-task. Previous researches~\cite{singh2023progprompt} adopt Python API \texttt{pick}(\textit{human\_arm},\textit{robot}) and use pre-defined primitives that the objects as well their states and positions are given. However, generating robot trajectories according to open-ended $\task_i$ can be difficult because $\task_i$ may be arbitrarily long-horizon or contextually sensitive due to the various environments. The symbolic concept \textit{human\_arm} as well as the motion related to it should be correctly grounded to the real world. The central problem investigated in this work is to generate a motion trajectory $\trajectory_i$ and each manipulation phase described by instruction $\task_i$, without the help of prior knowledge. 
\textcolor{black}{We use Python API $\action = \texttt{hold}(\textit{human\_arm},\textit{target\_pose}, \textit{robot\_hand})$, where the 6-DoF pose of target objects and collision detection are automatically predicted. Lastly, the model completes the motion generation according to the 6-DoF pose and collision detection results so as to implement the action execution.}
\subsection{Dataset Construction}
\label{sec:data_construction}

\paragraph{Crowd-Sourcing}
Our datasets comprise two kinds of multi-modal datasets: the video dataset and the image dataset with a greater variety of diverse scenes.

We collect video datasets capturing patient and elderly healthcare scenarios while wearing a head-mounted depth sensor, spatial microphone array, and a video camera\footnote{We use Azure Kinect DK~(\url{https://azure.microsoft.com/zh-cn/products/kinect-dk/}) that contains a depth sensor, spatial microphone array with a video camera, and orientation sensor as an all-in-one small device with software development kits (SDKs).}. The video was captured from a first-person perspective. This camera is capable of capturing synchronized and aligned RGB and depth images at a resolution of 1280x720, with a frame rate of 30Hz. During each data collection round, there are two participants, one participant plays the role of a service robot, equipped with head-mounted recording equipment, and the other plays the role of an elderly person or patient with a fragile body. 

Our video dataset includes a range of scenarios where humans require assistance from others, with varying levels of complexity from easy to difficult. Easy scenes involve individuals with no obstacles blocking the robot's path to the human. In more complex scenes, there may be obstacles or situations where the robot must rely solely on auditory cues to locate those in need. We collected data from 10 participants who all signed informed consent forms.

\textcolor{black}{Furthermore, our image dataset includes more scenarios, a portion of which is sourced from publicly available datasets. These challenging scenes cover a range of situations, including traffic accidents and natural disasters.}

\paragraph{Human Labeling}\label{sec:ALP-text2}
Then annotators annotated the data samples with task/sub-task/motion labels. Workers subsequently provided a double-checking by watching their performance. These labels enable future studies in ego-centric perception. 

\taskname dataset includes a rich set of information as follows: task/subtask labels, object mash, states, collisions, and 6-DoF pose of objects, \etc. 
As shown in Fig. \ref{fig:task}, we collected this multi-modal dataset with seven modalities including FPV (First-person view), TPV (Third-person view), Text (Communication text), Audio, IMU (Inertial measurement unit), Depth (Depth image), Pose). Additionally, the human pose in each frame of the dataset was labeled according to the COCO human pose style\footnote{\url{https://docs.ultralytics.com/datasets/pose/coco/}}. 


We utilize Amazon Mechanical Turk (AMT)\footnote{\url{https://www.mturk.com/}}, an online marketplace to hire remote workers to perform a crowd-sourcing survey. We created a survey in AMT that allows workers to label raw video data and do double-checking. We provide workers with comprehensive instructions and a set of well-defined examples to judge the labeling quality and modify those unsatisfied. During the labeling, we will show workers an interface with automatically labeled data for reference. The final results from three workers are combined using a majority vote. For more details, please refer to Appendix.

\paragraph{Quality Control}
We use a variety of quality control methods while labeling to ensure high-quality results. To achieve this, we release datasets to workers one by one, and at least two workers must label the same sample to check for any differences. Only workers with at least 500 successful hits and a high level of accuracy (more than 93\%) in their labeling history are allowed to label the data. After finishing the first round of initial labeling, we hire workers to double-check the results. We collect flags for any differences in multiple decisions made by several workers and manually double-check those samples with differences using a third-party worker. Additionally, we flag annotations from workers whose work appears inadequate and remove their results from the final collections.

\subsection{Dataset Analysis}\label{sec:data_analysis}
\paragraph{Video Dataset Statistics}
The detailed statistics of \taskname dataset are shown in the Appendix. The dataset includes raw data with seven modalities as well as various labels, such as human poses, task labels, and action labels. 

The dataset contains a total of 55 videos, 37157 RGB-D frames, and 13 motion categories. The total duration of recordings for all participants is over 25 minutes, with a total number of clips of 37157 × 4 (4 for the following video types including RGB video, depth image, audio, and skeleton).


\textcolor{black}{The dataset also contains image data covering about 5 classes with over 20 sub-classes. A portion of the dataset is from publicly available photo datasets, including FallDown\footnote{https://aistudio.baidu.com/datasetdetail/127208/0}, BowFire~\citep{chino2015bowfire}, and RAF-DB~\citep{li2017reliable}. Furthermore, we have acquired additional challenging scenes by crawling the internet. These challenging scenes contain various scenarios such as traffic accidents, natural disasters, and more. Tab.~\ref{tab:unlabeled-data-analysis} shows the data statistics of the image dataset.} 

\begin{table}[h]
\renewcommand\arraystretch{1.2}
\centering
\begin{tabular}{lcc}
\toprule
\textbf{Type} & \textbf{Source} & \textbf{Number} \\ 
\midrule
Fall Classfication&FallDown&7.7k \\
Fire Detection&BoWFire&0.2k \\
Fire Detection&FDDB&0.7k \\
Sentiment Analysis& RAB-DB&0.2k \\
Sentiment Analysis&MMAFEDB&2.5k \\
Disaster Detection&Self-Craw&1k \\
Injury Detection&Self-Craw&70 \\
Car Crash Detection&Self-Craw&35 \\
\bottomrule
\end{tabular}
\caption{The statistics of various types.}
\label{tab:unlabeled-data-analysis}
\vspace{-0.5cm}
\end{table}


\section{\model}\label{sec:approach}
\begin{figure*}[!t]
    \centering
    \includegraphics[width=1.0\linewidth]{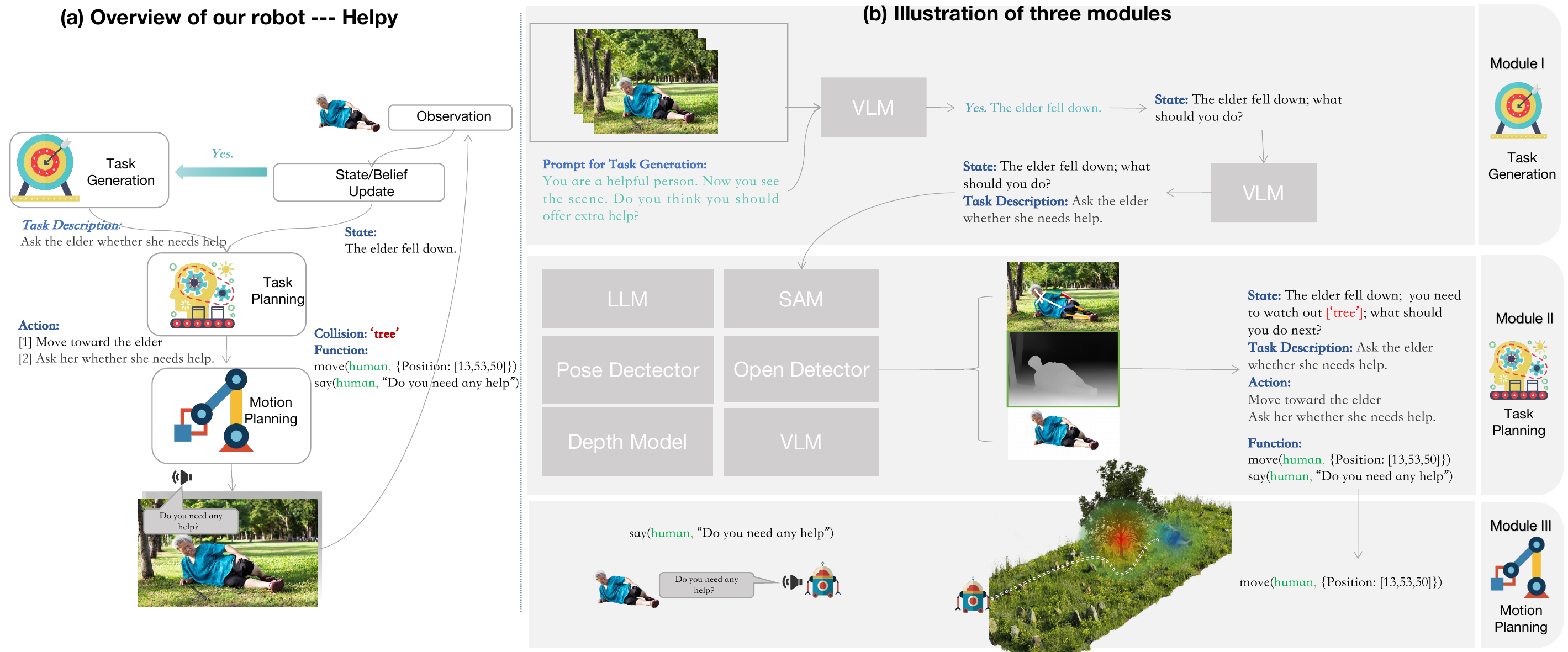}
   \caption{Overview of \model. As shown in the subfigure (a), \model includes five key components: 1) observation of the environment, 2) a state/belief of the environment scene, 3) a task generation module, 4) a task planner to find a plan for a generated task based on its state/belief and 5) motion planner to execute the plan. As shown in (b),
   given the first-person observation of the environment, the task generation module generates the task through prompt engineering.  The task generation module interacts with several vision toolkits to produce the plans as well as the augments for execution. The augments serve as objective functions for motion planners to synthesize trajectories for manipulation. Then, the motion planner executes a sequence of plans, such as generating a trajectory toward the target person and talking to the person.}
    \label{fig:model}
\end{figure*}

Fig. \ref{fig:model} illustrates the overview of \model. As shown in the subfigure (a), \model includes five key components to construct a loop. Subfigure (b) shows the three modules: task generation module, task planning module, and motion planning module.
We leverage this learning-free method due to the lack of a parallel dataset of the natural language instructions and the executable robot action sequences, especially in a open-ended real-world.

\subsection{Altruistic Task Generation}
We use the VLM~\citep{zhu2023minigpt} to guess the patient's intention and predict when help is needed so that it provides a helping service immediately.
The VLM with the model parameters as $\theta$ is trained using a maximum likelihood loss to estimate the probability of a sequence of tokens $\mathbf{y}$ based on the prompt $\mathbf{x}$, expressed as $\theta = \arg \max_{\theta} P(\mathbf{y} | \mathbf{x}; \theta)$.
The prompt $\mathbf{x}$ is as follows:

\textit{$<$Scene$>$ You are a helpful person. Now you see the scene. Do you think you should offer extra help?}

Subsequently, if the VLM judges that someone in the video needs help, then our approach utilizes grounded DINO~\citep{liu2023grounding} to detect humans in the given situation. For each detected human, we use the following prompt to inquire if the designated person (marked with a bounding box) is in need of assistance.

\textit{$<$Scene$>$ You are a helpful person. Does $<$designated person$>$ need assist in this picture?}

\subsection{Task Planning}
LLM\footnote{https://openai.com/blog/chatgpt} handles decomposition and sequencing of the task which is represented as $\task \to (\task_1, \task_2, \dots, \task_n)$). Prompting the LLM to generate sub-tasks given a text-based task description is an emerging topic~\cite{jansen2020visually,patel2021mapping,li2022pre,huang2022language,ahn2022can,huang2022inner}. 
\textcolor{black}{Here, each $\task_i$ is denoted as a Pythonic function with a text-based sub-task description. Each $\task_i$ is generated by the LLM. Given the ground truth as history, the task planner predicts one sub-task at a time. The types and corresponding Python functions will be dynamically adjusted based on the current state to ensure adaptability in a dynamically changing environment. }


\paragraph{Sub-task Representation}
\textcolor{black}{For plans and sub-tasks, we design the input prompt including exemplars, environmental status, and the high-level task to elicit an LLM to generate situated robot sub-tasks, conditioned on the $(\textit{state}, \textit{action}, \textit{function})$ triplet-style prompt, where \textit{state} represents the current environment state, which is determined by the current environment and historical \textit{state}, \textit{action} represents how the robot should act according to the current \textit{state}, and according to the current  \textit{state} and  \textit{action}, a Pythonic \textit{function} will be obtained to control the robot.}



We pre-define funtion names for the sub-task, including \texttt{say(human, context)}, \texttt{receive\_info()} and \texttt{pick(object, object\_pose, agent)}, \etc.
Unlike prior works, where objects \texttt{object} are text strings, we provided the grounded object that we map to mashes and point clouds using segmentation masks and the depth image.
\model provides the available objects in the environment in a real-time way instead of pre-defined.
Additionally, \model provides information about the environment and primitive actions to the LLM through prompts. 

\model also includes exemplars ---fully executable program plans for LLM's references. Exemplars demonstrate how to complete a given task description using available sub-tasks (actable actions) and objects in the physical environment.
These examples demonstrate the relationship between task name, given as the function handle, and task to split, splitting sub-tasks as well as the restrictions on actions and objects to involve.

\paragraph{Task Splitting}
Fig.~\ref{fig:model} (b) illustrates our prompt input which takes in all the information (first-person observations, \etc). 
The LLM then predicts the sequence of sub-tasks as well as the corresponding executable functions.
In the task \texttt{Task Description: Ask the elder whether she needs help}, a reasonable first sub-task that an LLM could generate is (\texttt{Move towards the elder} and \textit{move}(\textit{human}, \textit{Position}=[13,53,50]). Note that, \model uses three modalities: FPV, text, the depth images and implements  the speech-to-text model~\citep{radford2023robust} for speech recognition.

Sometimes, the first-person view of the robot might not have seen objectives for this sub-task (as shown in Fig. \ref{fig:task} (d)).
In this case, the VLM with the information from the real-time video encourages the robot to make reasonable assumptions to find the target. VLM restricts its views to only targets that are available in the current scene. Since our prompting scheme explicitly lists out the set of functions and objects available to the model, the generated plans typically contain actions a robot can take and objects available in the environment or history. 

For a precise perception of the first-person view, we utilize a process flow that incorporates state-of-art toolkits that specialize in different vision tasks, \ie, depth estimation~\cite{Ranftl2020}, pose detector for pose detection~\citep{xu2022vitpose}, open detector~\citep{liu2023grounding}, and SAM for image segmentation~\citep{kirillov2023segany}. These advanced visual models effectively provide precise scene information and accurately label the status and position of objects. 


\subsection{Motion Planning}
The generated plans are further tested on a physical mobile manipulator that is equipped with a Clearpath Ridgeback omnidirectional mobile base and two UR5 manipulators. We adopt the framework of~\cite{jiao2021consolidated,jiao2021efficient} to generate executable trajectories for both the mobile base and the manipulator. The robot executes the generated actions sequentially, which consist of tasks such as relocating objects, dialing the emergency call, navigating to a new location, and initiating a conversation. The 6-DoF pose for each action serves as the target input. 

Generated plans are executed on a physical robot system that executes each action command against the physical environment. The augments from the Python function serve as objective functions for motion planners to synthesize trajectories for manipulation. Then, the motion planner executes a
sequence of plans, such as generating a trajectory toward the target person with location coordinates of [13,53,50] and talking to the person: \textit{Do you need any help?}.
We use the text-to-speech model~\citep{kaur2023conventional}.

\section{Experiments}\label{sec:experiments}

\subsection{Setup}

\paragraph{Hyper-Parameters Setting}\label{app:hyperparameter}
\textcolor{black}{We use GPT-3.5 as the LLM model, MiniGPT-4~\citep{zhu2023minigpt} as the VLM model, and a series of vision models~\citep{Ranftl2020, xu2022vitpose, kirillov2023segany}. Parameter settings follow the original paper or the default setting. We use GroundedDINO~\citep{liu2023grounding} as an open-vocabulary object detection model to identify objects in the current scene.} We use 15-shot exemplars.
We utilize the labeled dataset to assess the performance of \model across various scenes. 

\paragraph{Evaluation Metrics}
Following \citet{singh2023progprompt}, we use three metrics: success rate (\sr), goal conditions recall (\gcr), and executability (\exec). The task-relevant goal-conditions recall judges whether the correct function name can be obtained according to the current environment state to execute the robot. The execution score (\exec) is mainly used to judge whether the generated function can be executed, and finally, the successful rate (\sr) judges the final state achieved the final goal with the generated function, divided by the number of task-specific goal-conditions; \sr$=1$ only if \gcr$=1$. 


\subsection{Results}

\paragraph{Results on Different Task Complexity}
Our dataset includes a range of scenarios with varying levels of complexity from easy to difficult. Easy samples involve individuals with no obstacles blocking the robot's path to the human. In complex samples, there may be obstacles or situations where the robot must rely solely on auditory cues to locate those in need. We quantitatively present the results of \model in scenarios with different complexity in Tab.~\ref{tab:difficulty}. 
The easiest samples (Row \textit{No harm}) score is 0.6476 instead of 1.0 in Tab.~\ref{tab:difficulty}, suggesting that the current VLM model capability cannot stably and accurately judge whether the current scenario needs help. To note, the random baseline obtains the \sr of 0.091.

\begin{figure}[!t]
    \centering
    \includegraphics[width=1\linewidth]{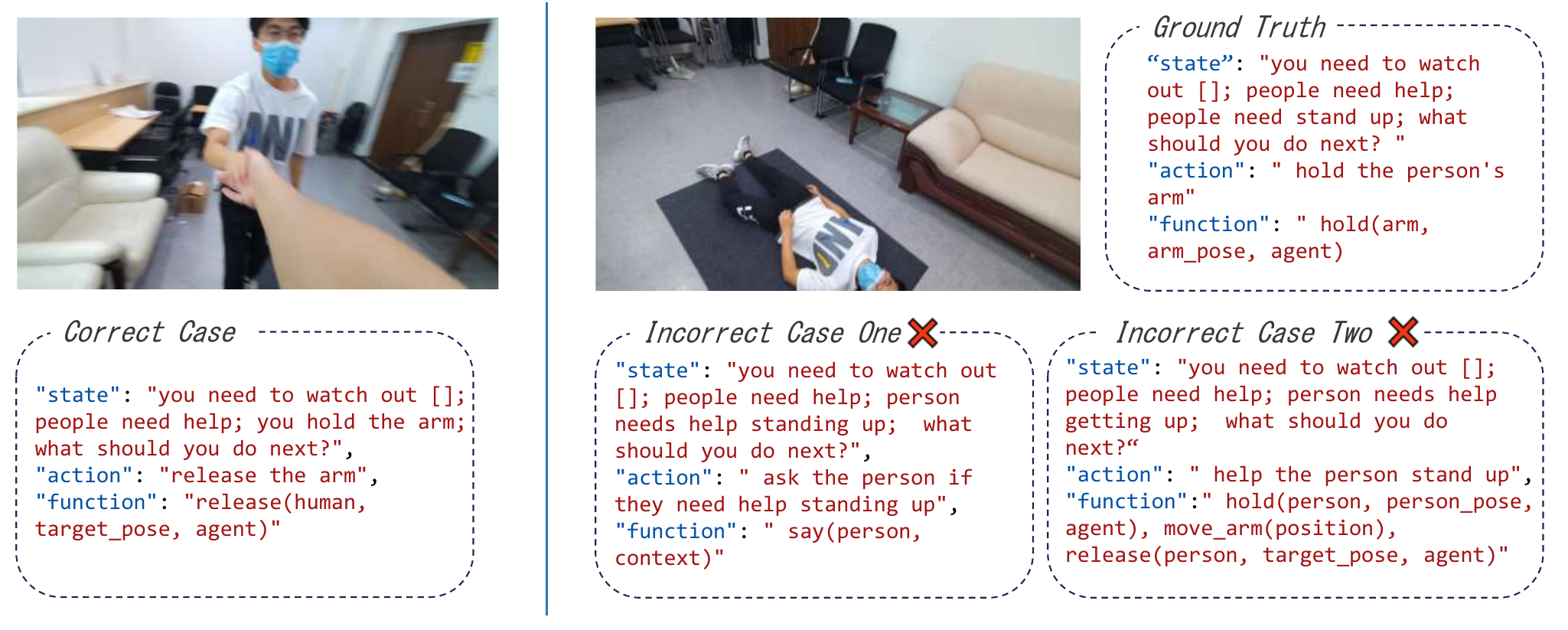}
    \caption{Case study of \model.}
    \label{fig:case}
\end{figure}

See Tab.~\ref{tab:difficulty} and Fig.~\ref{fig:case} (a) for success examples. \model effectively assesses the present condition and provides the appropriate handler function for the release of the human arm. However,  \model is far from satisfaction. Fig.~\ref{fig:case} (b) demonstrates a typical failure case. When \model receives the information that assistance is required for the person to stand up, it often faces difficulty in accurately determining the appropriate functions for the next step. In the first incorrect case, an incorrect execution function is returned based on the current state, mistakenly assuming that continuous questioning is necessary. On the other hand, in incorrect case two, multiple functions are returned, which hinders the model's ability to carry out the tasks effectively.
This failure case reveals the importance of appropriately modeling scene information, which highlights the major challenge of our proposed task and is calling for future research for better VL models.

\begin{table}[ht]
\centering
\newcommand\robotasktext[1]{\footnotesize\textit{#1}}
\begin{tabular}{l@{}cccc}
    \toprule
    \textbf{Complexity} \quad\quad & \textbf{Num} & \textbf{\sr} & \textbf{\exec} & \textbf{\gcr}\\
    \toprule
    \multirow{1}{*}{\robotasktext{No harm}}
     & 618 & 0.6474 & 0.6474 & 0.6474 \\
    \midrule
    \multirow{1}{*}{\robotasktext{Easy}}
     & 570 & 0.3012 & 0.5569 & 0.3327 \\
    \midrule
    \robotasktext{Medium}
     & 747 & 0.4297 & 0.6760 & 0.4564 \\
     \midrule
    \robotasktext{Hard}
     & 456 & 0.1233 & 0.4943 & 0.1299 \\
    \midrule
    \robotasktext{Total}
     & 2391 & 0.3969 & 0.6055 & 0.4140 \\
    \bottomrule
\end{tabular}
\caption{Results on the physical robot. From top to bottom, the task complexity gradually increases.}
\label{tab:difficulty}
\end{table}



\paragraph{Results on Different Task Types}

The results of different task types are presented in Tab.~\ref{tab:main_result}. 
It is also prone to random failures such as grasping slips. The real world introduces randomness that makes it difficult to compare systems quantitatively. As demonstrated in Tab.~\ref{tab:main_result}, \model performs better in basic tasks, such as determining if anyone needs help, waiting for a human response, holding and releasing the human's arm. These tasks involve minimal compositional steps as the model only needs to assess the initial and final states to determine the necessary action. 
\textcolor{black}{However, more complex tasks require a thorough reasoning process. For instance, after receiving a human response, \model needs to analyze the collected information to determine the subsequent actions. This cannot be achieved through a single fixed function. The success of the ``\textit{move to the human}'' task relies on the depth backbone, although there may still be notable discrepancies between the depth prediction model and the real-life scenario. Consequently, \model performs poorly in the challenging task of picking up the obstacle. This task necessitates the identification of obstacles, followed by determining their location. The model then executes the pick function based on the 6-DoF pose of the obstacle.}

\begin{table}[ht]
\centering
\newcommand\robotasktext[1]{\footnotesize\textit{#1}}
\begin{tabular}{l@{}cccc}
    \toprule
     \textbf{Task Description} & $\mathbf{Num}$ & $\mathbf{\sr}$ & $\mathbf{\exec}$ & $\mathbf{GCR}$ \\
    \toprule
    \robotasktext{Ask if help needed}
     & 254 & 0.6220 & 0.8421 &  0.6299 \\
    \midrule
    \robotasktext{Move to the human}
     & 138 & 0.0745 & 0.3428 & 0.3205 \\
    \midrule
    \robotasktext{Wait human response}
    & 38 & 0.5789 & 0.8421 &  0.5789\\
    \midrule
    \robotasktext{Receive human response}
   & 185 & 0.0424 & 0.6540 & 0.0434 \\
     \midrule
    \robotasktext{Hold the human's arm}
   & 196 & 0.0724 & 0.7806 & 0.1378 \\
    \midrule
    \robotasktext{Release the human's arm}
   & 237 &  0.2954 & 0.6456 &  0.4261 \\
    \midrule
    \robotasktext{Pick up chair}
     & 64 & 0.0356 & 0.5468 &  0.2342 \\
    \midrule
    \robotasktext{Total}
     & 1112 & 0.2559 & 0.6791 & 0.3392  \\
    \bottomrule
\end{tabular}
\caption{Main result for different sub-tasks. $\mathbf{Num}$ denotes the number of sub-tasks.}
\label{tab:main_result}
\end{table}

\section{Related Work}\label{sec:related_work}
\subsection{LLM as Task Planner}
Designing prompts to produce desired texts that are useful for robot task planning is a new area of research~\cite{jansen2020visually,patel2021mapping,li2022pre,huang2022language,ahn2022can,huang2022inner}. Designing a prompt is challenging because there is a lack of natural language instructions paired with executable plans~\cite{liu2023pre}. To develop a prompt for task plan prediction, two steps are needed to complete: creating a prompting function and an answer extraction strategy~\cite{liu2023pre}. The prompting function converts the environmental state $\state$ into a textual prompt. For the answer search, LLM generates an output directly or scores a predefined set of options.

\citet{huang2022language} generates open-domain plans using LLMs by first open-ended task plan generation without any environment interaction, followed by action matching and ultimately testing the feasibility of the matched actions.
Due to a lack of environmental interaction and post-testing, generated action doesn't ensure that the action is admissible in the current situation. 
\citet{huang2022inner} propose \textsc{InnerMonologue} to leverage environment feedback and state monitoring. However, this problem is not going away. The proposed actions presented by LLM planners involved objects that were not always present in the scene. SayCan~\citep{ahn2022can} not only uses prompting to generate feasible planning steps, but also re-scoring matched admissible actions using a value function.  
Socratic models~\citep{zeng2022socratic} use code-completion to generate robot plans. 
\citet{singh2023progprompt} also shows that a programming language-inspired prompt (ProgPrompt) can inform the LLM of both situated environment state and available robot actions.
Unlike \citet{zeng2022socratic}, ProgPrompt defines a more sophisticated prompt design that includes import statements to model robot capabilities, comments, and assertions to track the execution state.
For ensuring output compatibility to robot actions, \model uses real-time video streaming and processes it with VLM so as to monitor the real-time environment feedback.

\subsection{Combining Task Planning and Motion Planning}
For high-level task planning, early studies work in robotics use search in a pre-defined domain~\cite{fikes1971strips,garrett2020pddlstream,jiang2018task}. Due to the large branching factors in the real world, scaling in environments with many feasible actions and objects can be challenging~\cite{puig2018virtualhome,shridhar2020alfred}. 
To alleviate this problem, heuristics, and prior knowledge are often leveraged to guide the search-based planning~\cite{baier2009heuristic,hoffmann2001ff,helmert2006fast,bryce2007tutorial}. 
Recent works~\cite{zhu2021hierarchical,sharma2022skill} have explored learning-based task \& motion planning. 
Concurrent to our work, \citet{huang2023voxposer} introduces a robot manipulation framework by composing task planning and motion planning and generating 3D value maps grounded in observation space.


We go beyond VoxPoser by extending programming languages with real-time environment feedback, collision information, 6-DoF, \etc.
Thus, an entire, executable plan program is performed by allowing this additional information as arguments used to synthesize trajectories for manipulation tasks.
More importantly, \model completed the final part of the loop by designing a starting mechanism for ``task generation-task planning-motion planning'' loop.

\section{Conclusion and Further Work}
The global market for service robots is expected to grow significantly in the coming years, as service robots have the potential to revolutionize the lives of many people, especially those who are elderly, disabled, or patients. 
For the community's further studying and exploration, we introduce a new challenging task – \taskname Challenge for healthcare scenarios without human intervention commands, holistically evaluating the autonomous capabilities of robots, including task generation, task planning, and motion planning. 
We also provide a potential approach with respect to this challenge. 

Although \model has shown promising results, it still falls short of expectations, particularly in complex scenarios. Error propagation may occur. Moving forward, what is the best way to complete the whole loop? A promising extension could be to unify LLM and VLM to build powerful multi-modal foundation models to eliminate the gap between action function and language instructions. 



\bibliography{aaai24}

\begin{thebibliography}{43}
\providecommand{\natexlab}[1]{#1}

\bibitem[{Ahn et~al.(2022)Ahn, Brohan, Brown, Chebotar, Cortes, David, Finn, Fu, Gopalakrishnan, Hausman et~al.}]{ahn2022can}
Ahn, M.; Brohan, A.; Brown, N.; Chebotar, Y.; Cortes, O.; David, B.; Finn, C.; Fu, C.; Gopalakrishnan, K.; Hausman, K.; et~al. 2022.
\newblock Do as i can, not as i say: Grounding language in robotic affordances.
\newblock \emph{arXiv preprint arXiv:2204.01691}.

\bibitem[{Baier, Bacchus, and McIlraith(2009)}]{baier2009heuristic}
Baier, J.~A.; Bacchus, F.; and McIlraith, S.~A. 2009.
\newblock A heuristic search approach to planning with temporally extended preferences.
\newblock \emph{Artificial Intelligence}, 173(5-6): 593--618.

\bibitem[{Bekey(2005)}]{bekey2005autonomous}
Bekey, G.~A. 2005.
\newblock \emph{Autonomous robots: from biological inspiration to implementation and control}.
\newblock MIT press.

\bibitem[{Bollini et~al.(2013)Bollini, Tellex, Thompson, Roy, and Rus}]{bollini2013interpreting}
Bollini, M.; Tellex, S.; Thompson, T.; Roy, N.; and Rus, D. 2013.
\newblock Interpreting and executing recipes with a cooking robot.
\newblock In \emph{Experimental Robotics: The 13th International Symposium on Experimental Robotics}, 481--495. Springer.

\bibitem[{Bryce and Kambhampati(2007)}]{bryce2007tutorial}
Bryce, D.; and Kambhampati, S. 2007.
\newblock A tutorial on planning graph based reachability heuristics.
\newblock \emph{AI Magazine}, 28(1): 47--47.

\bibitem[{Chino et~al.(2015)Chino, Avalhais, Rodrigues, and Traina}]{chino2015bowfire}
Chino, D.~Y.; Avalhais, L.~P.; Rodrigues, J.~F.; and Traina, A.~J. 2015.
\newblock Bowfire: detection of fire in still images by integrating pixel color and texture analysis.
\newblock In \emph{2015 28th SIBGRAPI conference on graphics, patterns and images}, 95--102. IEEE.

\bibitem[{Fikes and Nilsson(1971)}]{fikes1971strips}
Fikes, R.~E.; and Nilsson, N.~J. 1971.
\newblock STRIPS: A new approach to the application of theorem proving to problem solving.
\newblock \emph{Artificial intelligence}, 2(3-4): 189--208.

\bibitem[{Garrett, Lozano-P{\'e}rez, and Kaelbling(2020)}]{garrett2020pddlstream}
Garrett, C.~R.; Lozano-P{\'e}rez, T.; and Kaelbling, L.~P. 2020.
\newblock Pddlstream: Integrating symbolic planners and blackbox samplers via optimistic adaptive planning.
\newblock In \emph{Proceedings of the International Conference on Automated Planning and Scheduling}, volume~30, 440--448.

\bibitem[{Helmert(2006)}]{helmert2006fast}
Helmert, M. 2006.
\newblock The fast downward planning system.
\newblock \emph{Journal of Artificial Intelligence Research}, 26: 191--246.

\bibitem[{Hoffmann(2001)}]{hoffmann2001ff}
Hoffmann, J. 2001.
\newblock FF: The fast-forward planning system.
\newblock \emph{AI magazine}, 22(3): 57--57.

\bibitem[{Holland et~al.(2021)Holland, Kingston, McCarthy, Armstrong, O’Dwyer, Merz, and McConnell}]{holland2021service}
Holland, J.; Kingston, L.; McCarthy, C.; Armstrong, E.; O’Dwyer, P.; Merz, F.; and McConnell, M. 2021.
\newblock Service robots in the healthcare sector.
\newblock \emph{Robotics}, 10(1): 47.

\bibitem[{Huang et~al.(2022{\natexlab{a}})Huang, Abbeel, Pathak, and Mordatch}]{huang2022language}
Huang, W.; Abbeel, P.; Pathak, D.; and Mordatch, I. 2022{\natexlab{a}}.
\newblock Language models as zero-shot planners: Extracting actionable knowledge for embodied agents.
\newblock In \emph{International Conference on Machine Learning}, 9118--9147. PMLR.

\bibitem[{Huang et~al.(2023)Huang, Wang, Zhang, Li, Wu, and Fei-Fei}]{huang2023voxposer}
Huang, W.; Wang, C.; Zhang, R.; Li, Y.; Wu, J.; and Fei-Fei, L. 2023.
\newblock VoxPoser: Composable 3D Value Maps for Robotic Manipulation with Language Models.
\newblock \emph{arXiv preprint arXiv:2307.05973}.

\bibitem[{Huang et~al.(2022{\natexlab{b}})Huang, Xia, Xiao, Chan, Liang, Florence, Zeng, Tompson, Mordatch, Chebotar et~al.}]{huang2022inner}
Huang, W.; Xia, F.; Xiao, T.; Chan, H.; Liang, J.; Florence, P.; Zeng, A.; Tompson, J.; Mordatch, I.; Chebotar, Y.; et~al. 2022{\natexlab{b}}.
\newblock Inner Monologue: Embodied Reasoning through Planning with Language Models.
\newblock In \emph{6th Annual Conference on Robot Learning}.

\bibitem[{Jansen(2020)}]{jansen2020visually}
Jansen, P. 2020.
\newblock Visually-Grounded Planning without Vision: Language Models Infer Detailed Plans from High-level Instructions.
\newblock In \emph{Findings of the Association for Computational Linguistics: EMNLP 2020}, 4412--4417.

\bibitem[{Jiang et~al.(2018)Jiang, Zhang, Khandelwal, and Stone}]{jiang2018task}
Jiang, Y.; Zhang, S.; Khandelwal, P.; and Stone, P. 2018.
\newblock Task Planning in Robotics: an Empirical Comparison of PDDL-based and ASP-based Systems.
\newblock \emph{arXiv preprint arXiv:1804.08229}.

\bibitem[{Jiao et~al.(2021{\natexlab{a}})Jiao, Zhang, Jiang, Han, Zhu, Zhu, and Liu}]{jiao2021consolidated}
Jiao, Z.; Zhang, Z.; Jiang, X.; Han, D.; Zhu, S.-C.; Zhu, Y.; and Liu, H. 2021{\natexlab{a}}.
\newblock Consolidating Kinematic Models to Promote Coordinated Mobile Manipulations.

\bibitem[{Jiao et~al.(2021{\natexlab{b}})Jiao, Zhang, Wang, Han, Zhu, Zhu, and Liu}]{jiao2021efficient}
Jiao, Z.; Zhang, Z.; Wang, W.; Han, D.; Zhu, S.-C.; Zhu, Y.; and Liu, H. 2021{\natexlab{b}}.
\newblock Efficient Task Planning for Mobile Manipulation: a Virtual Kinematic Chain Perspective.

\bibitem[{Kaur and Singh(2023)}]{kaur2023conventional}
Kaur, N.; and Singh, P. 2023.
\newblock Conventional and contemporary approaches used in text to speech synthesis: A review.
\newblock \emph{Artificial Intelligence Review}, 56(7): 5837--5880.

\bibitem[{Kirillov et~al.(2023)Kirillov, Mintun, Ravi, Mao, Rolland, Gustafson, Xiao, Whitehead, Berg, Lo, Doll{\'a}r, and Girshick}]{kirillov2023segany}
Kirillov, A.; Mintun, E.; Ravi, N.; Mao, H.; Rolland, C.; Gustafson, L.; Xiao, T.; Whitehead, S.; Berg, A.~C.; Lo, W.-Y.; Doll{\'a}r, P.; and Girshick, R. 2023.
\newblock Segment Anything.
\newblock \emph{arXiv:2304.02643}.

\bibitem[{Kollar et~al.(2010)Kollar, Tellex, Roy, and Roy}]{kollar2010toward}
Kollar, T.; Tellex, S.; Roy, D.; and Roy, N. 2010.
\newblock Toward understanding natural language directions.
\newblock In \emph{2010 5th ACM/IEEE International Conference on Human-Robot Interaction (HRI)}, 259--266. IEEE.

\bibitem[{Li, Deng, and Du(2017)}]{li2017reliable}
Li, S.; Deng, W.; and Du, J. 2017.
\newblock Reliable crowdsourcing and deep locality-preserving learning for expression recognition in the wild.
\newblock In \emph{Proceedings of the IEEE conference on computer vision and pattern recognition}, 2852--2861.

\bibitem[{Li et~al.(2022)Li, Puig, Paxton, Du, Wang, Fan, Chen, Huang, Aky{\"u}rek, Anandkumar et~al.}]{li2022pre}
Li, S.; Puig, X.; Paxton, C.; Du, Y.; Wang, C.; Fan, L.; Chen, T.; Huang, D.-A.; Aky{\"u}rek, E.; Anandkumar, A.; et~al. 2022.
\newblock Pre-trained language models for interactive decision-making.
\newblock \emph{Advances in Neural Information Processing Systems}, 35: 31199--31212.

\bibitem[{Li, Liu, and Rehg(2023)}]{li2023eye}
Li, Y.; Liu, M.; and Rehg, J.~M. 2023.
\newblock In the Eye of the Beholder: Gaze and Actions in First Person Video.
\newblock \emph{IEEE Transactions on Pattern Analysis \& Machine Intelligence}, 45(06): 6731--6747.

\bibitem[{Liu et~al.(2023{\natexlab{a}})Liu, Yuan, Fu, Jiang, Hayashi, and Neubig}]{liu2023pre}
Liu, P.; Yuan, W.; Fu, J.; Jiang, Z.; Hayashi, H.; and Neubig, G. 2023{\natexlab{a}}.
\newblock Pre-train, prompt, and predict: A systematic survey of prompting methods in natural language processing.
\newblock \emph{ACM Computing Surveys}, 55(9): 1--35.

\bibitem[{Liu et~al.(2023{\natexlab{b}})Liu, Zeng, Ren, Li, Zhang, Yang, Li, Yang, Su, Zhu et~al.}]{liu2023grounding}
Liu, S.; Zeng, Z.; Ren, T.; Li, F.; Zhang, H.; Yang, J.; Li, C.; Yang, J.; Su, H.; Zhu, J.; et~al. 2023{\natexlab{b}}.
\newblock Grounding dino: Marrying dino with grounded pre-training for open-set object detection.
\newblock \emph{arXiv preprint arXiv:2303.05499}.

\bibitem[{Liu et~al.(2022)Liu, Liu, Jiang, Lyu, Wan, Shen, Liang, Fu, Wang, and Hoi4d}]{liu20224d}
Liu, Y.; Liu, Y.; Jiang, C.; Lyu, K.; Wan, W.; Shen, H.; Liang, B.; Fu, Z.; Wang, H.; and Hoi4d, L.~Y. 2022.
\newblock A 4d egocentric dataset for category-level humanobject interaction.
\newblock In \emph{Proceedings of the IEEE/CVF Conference on Computer Vision and Pattern Recognition}, 21013--21022.

\bibitem[{Lu et~al.(2020)Lu, Wirtz, Kunz, Paluch, Gruber, Martins, and Patterson}]{lu2020service}
Lu, V.~N.; Wirtz, J.; Kunz, W.~H.; Paluch, S.; Gruber, T.; Martins, A.; and Patterson, P.~G. 2020.
\newblock Service robots, customers and service employees: what can we learn from the academic literature and where are the gaps?
\newblock \emph{Journal of Service Theory and Practice}, 30(3): 361--391.

\bibitem[{Patel and Pavlick(2021)}]{patel2021mapping}
Patel, R.; and Pavlick, E. 2021.
\newblock Mapping language models to grounded conceptual spaces.
\newblock In \emph{International Conference on Learning Representations}.

\bibitem[{Puig et~al.(2018)Puig, Ra, Boben, Li, Wang, Fidler, and Torralba}]{puig2018virtualhome}
Puig, X.; Ra, K.; Boben, M.; Li, J.; Wang, T.; Fidler, S.; and Torralba, A. 2018.
\newblock Virtualhome: Simulating household activities via programs.
\newblock In \emph{Proceedings of the IEEE Conference on Computer Vision and Pattern Recognition}, 8494--8502.

\bibitem[{Puig et~al.(2020)Puig, Shu, Li, Wang, Liao, Tenenbaum, Fidler, and Torralba}]{puig2020watch}
Puig, X.; Shu, T.; Li, S.; Wang, Z.; Liao, Y.-H.; Tenenbaum, J.~B.; Fidler, S.; and Torralba, A. 2020.
\newblock Watch-And-Help: A Challenge for Social Perception and Human-AI Collaboration.
\newblock In \emph{International Conference on Learning Representations}.

\bibitem[{Radford et~al.(2023)Radford, Kim, Xu, Brockman, McLeavey, and Sutskever}]{radford2023robust}
Radford, A.; Kim, J.~W.; Xu, T.; Brockman, G.; McLeavey, C.; and Sutskever, I. 2023.
\newblock Robust speech recognition via large-scale weak supervision.
\newblock In \emph{International Conference on Machine Learning}, 28492--28518. PMLR.

\bibitem[{Ranftl et~al.(2020)Ranftl, Lasinger, Hafner, Schindler, and Koltun}]{Ranftl2020}
Ranftl, R.; Lasinger, K.; Hafner, D.; Schindler, K.; and Koltun, V. 2020.
\newblock Towards Robust Monocular Depth Estimation: Mixing Datasets for Zero-shot Cross-dataset Transfer.
\newblock \emph{IEEE Transactions on Pattern Analysis and Machine Intelligence (TPAMI)}.

\bibitem[{Schraft and Schmierer(2000)}]{schraft2000service}
Schraft, R.~D.; and Schmierer, G. 2000.
\newblock \emph{Service robots}.
\newblock CRC Press.

\bibitem[{Sharma, Torralba, and Andreas(2022)}]{sharma2022skill}
Sharma, P.; Torralba, A.; and Andreas, J. 2022.
\newblock Skill Induction and Planning with Latent Language.
\newblock In \emph{Proceedings of the 60th Annual Meeting of the Association for Computational Linguistics (Volume 1: Long Papers)}, 1713--1726.

\bibitem[{Shridhar et~al.(2020)Shridhar, Thomason, Gordon, Bisk, Han, Mottaghi, Zettlemoyer, and Fox}]{shridhar2020alfred}
Shridhar, M.; Thomason, J.; Gordon, D.; Bisk, Y.; Han, W.; Mottaghi, R.; Zettlemoyer, L.; and Fox, D. 2020.
\newblock Alfred: A benchmark for interpreting grounded instructions for everyday tasks.
\newblock In \emph{Proceedings of the IEEE/CVF conference on computer vision and pattern recognition}, 10740--10749.

\bibitem[{Singh et~al.(2023)Singh, Blukis, Mousavian, Goyal, Xu, Tremblay, Fox, Thomason, and Garg}]{singh2023progprompt}
Singh, I.; Blukis, V.; Mousavian, A.; Goyal, A.; Xu, D.; Tremblay, J.; Fox, D.; Thomason, J.; and Garg, A. 2023.
\newblock Progprompt: Generating situated robot task plans using large language models.
\newblock In \emph{2023 IEEE International Conference on Robotics and Automation (ICRA)}, 11523--11530. IEEE.

\bibitem[{Tellex et~al.(2020)Tellex, Gopalan, Kress-Gazit, and Matuszek}]{tellex2020robots}
Tellex, S.; Gopalan, N.; Kress-Gazit, H.; and Matuszek, C. 2020.
\newblock Robots that use language.
\newblock \emph{Annual Review of Control, Robotics, and Autonomous Systems}, 3: 25--55.

\bibitem[{Tellex et~al.(2011)Tellex, Kollar, Dickerson, Walter, Banerjee, Teller, and Roy}]{tellex2011understanding}
Tellex, S.; Kollar, T.; Dickerson, S.; Walter, M.; Banerjee, A.; Teller, S.; and Roy, N. 2011.
\newblock Understanding natural language commands for robotic navigation and mobile manipulation.
\newblock In \emph{Proceedings of the AAAI Conference on Artificial Intelligence}, volume~25, 1507--1514.

\bibitem[{Xu et~al.(2022)Xu, Zhang, Zhang, and Tao}]{xu2022vitpose}
Xu, Y.; Zhang, J.; Zhang, Q.; and Tao, D. 2022.
\newblock Vi{TP}ose: Simple Vision Transformer Baselines for Human Pose Estimation.
\newblock In \emph{Advances in Neural Information Processing Systems}.

\bibitem[{Zeng et~al.(2022)Zeng, Attarian, Choromanski, Wong, Welker, Tombari, Purohit, Ryoo, Sindhwani, Lee et~al.}]{zeng2022socratic}
Zeng, A.; Attarian, M.; Choromanski, K.~M.; Wong, A.; Welker, S.; Tombari, F.; Purohit, A.; Ryoo, M.~S.; Sindhwani, V.; Lee, J.; et~al. 2022.
\newblock Socratic Models: Composing Zero-Shot Multimodal Reasoning with Language.
\newblock In \emph{The Eleventh International Conference on Learning Representations}.

\bibitem[{Zhu et~al.(2023)Zhu, Chen, Shen, Li, and Elhoseiny}]{zhu2023minigpt}
Zhu, D.; Chen, J.; Shen, X.; Li, X.; and Elhoseiny, M. 2023.
\newblock Minigpt-4: Enhancing vision-language understanding with advanced large language models.
\newblock \emph{arXiv preprint arXiv:2304.10592}.

\bibitem[{Zhu et~al.(2021)Zhu, Tremblay, Birchfield, and Zhu}]{zhu2021hierarchical}
Zhu, Y.; Tremblay, J.; Birchfield, S.; and Zhu, Y. 2021.
\newblock Hierarchical planning for long-horizon manipulation with geometric and symbolic scene graphs.
\newblock In \emph{2021 IEEE International Conference on Robotics and Automation (ICRA)}, 6541--6548. IEEE.

\end{thebibliography}

\appendix
\section{Comparison with Previous Work}
We conduct the human evaluation to validate the previous SOTA (namely, Progprompt~\citep{singh2023progprompt}) and \model.
We convene 10 workers to score the results of the ProgPrompt and our \model model on 10 samples randomly selected from our \taskname dataset to judge whether the functions generated by the two systems can complete the corresponding tasks in reality. The full score of each sample is 10, and the total score for 10 samples is 100. The scores from low to high indicate the possibility of the model completing the corresponding task. The higher, the better. A score of 0 means that the model is completely unable to generate an effective function to complete the task and a score of 10 means that the model can generate an effective function to complete the current task. 

We obtain an average score of 29 for ProgPrompt, and 73 for \model. It can be observed that the ProgPrompt cannot adapt to changes in the environment, such as monitoring the open-ended environment status. Meanwhile, the \model model effectively compensates for this issue. 

\section{Ablation Study}
As illustrated in the ablation study experiment (tab. \ref{tab:ablation_study}), when the \texttt{action} attribute was removed, due to the lack of clear instructions, the success rate of completing the task decreased. After removing the history state, the SR score and GCR score declined to varying degrees. 

\begin{table}[ht]
\centering
\newcommand\robotasktext[1]{\footnotesize\textit{#1}}
\begin{tabular}{lcccc}
    \toprule
     \textbf{Approach} & $\mathbf{Num}$ & $\mathbf{\sr}$ & $\mathbf{\exec}$ & $\mathbf{GCR}$ \\
    \toprule
    \robotasktext{\model}
    & 747 & 0.4297 & 0.6760 & 0.4564 \\
    \midrule
    $\quad$\robotasktext{-action}
      & 747 & 0.3748 & 0.7215 & 0.3962 \\
    \midrule
    $\quad$\robotasktext{-history}
  & 747 & 0.3739 & 0.6876 & 0.3860 \\
    \midrule
\end{tabular}
\caption{Ablation study.}
\label{tab:ablation_study}
\end{table}

\section{Analysis of Different Prompts}\label{app:robust}
We conduct experiments to analyze the effects of different prompts on our model. The analysis is conducted on the task generation module. Our experiments show that our method is robust in terms of the wording and phrasing of prompts. We used 100 randomly selected samples for our experimental data, and the detailed results are shown in Table \ref{table:robust_result}. We observed that the variance is very small, suggesting that changes in the textual prompts due to different wording and phrasing do not have a significant impact on our model's performance. This demonstrates the robustness of our approach.

\begin{table*}[!tb]
\centering
\resizebox{0.98\textwidth}{!}{
\begin{tabular}{ccc}
\toprule
 No. & Template & \sr \\
 \midrule
 1 & \leftline{Please recognize the obstruction of \{\}, and the human is \{\}, what's your next move?} 
 &  0.75 \\
 2 & \leftline{the obstacle includes \{\}, the current circumstance is \{\}.What's the next step you should take?} &  0.74\\
 3 & \leftline{Be careful with the \{\}!, people may require \{\}, What should you do now?.}
 &  0.71 \\
 4 & \leftline{you need to watch out \{\}; people need \{\}, what should you do next?}
 & 0.73\\

 Average &  &  0.7325\\
 Variance &  & \textit{0.291\%} \\
\bottomrule
\end{tabular}
}

\caption{Results of task generation module for different prompts. }
\label{table:robust_result}
\end{table*}

\section{Function List}
We list the functions as follows:
\begin{itemize}
    \item \texttt{pick(object,objectpose,agent)}

    \item \texttt{place(object,target\_pose,agent)}

    \item \texttt{hold(object,objectpose,agent)}

    \item \texttt{release(object,target\_pose,agent)}

    \item \texttt{rotation(angle)}

    \item \texttt{wait(second)}

    \item \texttt{phone\_dial(number)}

    \item \texttt{say(object, context)}

    \item \texttt{receive\_info()}

    \item \texttt{stop()}

    \item \texttt{move(target\_position)}

    \item \texttt{move\_base(target\_position)}

    \item \texttt{move\_arm(target\_position)}
\end{itemize}

\section{Prompt Design}


\begin{lstlisting}[language=Python, basicstyle=\fontsize{9}{10}\selectfont\fontfamily{pcr}\selectfont]
You are an active help robot. When someone needs help, you need to go to the side of the person and ask what help you need. If the person needs help at this time, you need to ask clearly what you need help with. And give help. After the help is over, you need to ask again if the person needs other help. If the person answers that you do not need it, please go back to the original place and turn it off. But when the person does not get an answer from the other party, please dial 120 immediately and report the current situation.
You now have the following function calls: 
pick(object, object_pose, agent)
place(object, target_pose, agent)
hold(object, object_pose, agent)
release(object, target_pose, agent)
rotation(angle)
wait(second)
phone_dial(number)
say(object, context)
receive_info()
stop()
move(target_position)
move_base(target_position)
move_arm(target_position)


Introduction:
When there is an obstacle in front of you that prevents you from moving forward, you need to call the pick(object, objectpose, agent) function, and then call place(object, target_pose, agent).
When you need to move to the position of the helper, you need to call move(target_position)
When you have reached the helper, you need to call say(object, context) to ask if you need help.
After you have finished speaking, you need to wait for the reply from some else using the wait(second) function. 
When you receive a reply from someone, you need to call the receive_info() function to receive information until the person finishes speaking.
When you need to help him stand up, you need to use function hold(object, object_pose, agent) to grab the person arm 
After moving the person arm to a specific position, you need to use the release(object, target_pose, agent) function to release the arm.
When all help is over, call the stop() function to end the execution.

Please use the following format to find the pattern and output the next action and function.

state: ""
action: ""
function: ""

state: "you need to watch out []; people need help; what should you do next?"
action: "ask him whether he needs help"
function: "say(human, context)"

state: "you need to watch out []; people need help; people are far from you; what should you do next?"
action: "move to the people"
function: "move_base(human_position)"

state: "you need to watch out []; people need help; people behind the door; what should you do next?"
action: "hold the door handle"
function: "hold(handle, handle_pose, agent)"

state: "you need to watch out []; people need help; people behind the door; you hold the door handle; what should you do next?"
action: "rotate the door handle"
function: "rotation(handle, angle)"

state: "you need to watch out []; people need help; people behind the door; you rotate the door handle; what should you do next?"
action: "open the door"
function: "release(handle, target_pose, agent)"

state: "you need to watch out []; people need help; wait people to reply me; what should you do next?"
action:  "wait for a response"
function:  "wait(second)"


state: "you need to watch out []; people need help; you did not receive a response; what should you do next?"
action:  "call 120"
function:  "phone_dial(number)"

state: "you need to watch out []; people need help; people reply me; what should you do next?"
action: "receive response"
function: "receive_info()"

state:"you need to watch out [chair, ]; people need help; people need stand up; what should you do next?"
action: "remove the chair"
function: "pick(chair, chair_pose, agent)"

state:"you need to watch out [chair, ]; people need help; people need stand up; you pick the chair; what should you do next?"
action: "place the chair"
function: "place(chair, target_pose, agent)"

state:"you need to watch out []; people need help; people need stand up; what should you do next?"
action: "hold the arm"
function: "hold(arm, arm_pose, agent)"

state:"you need to watch out []; people need help; you hold the arm; what should you do next?"
action: "release the arm"
function: "release(human, target_pose, agent)"

state: "you need to watch out []; people does not need help; what should you do next?"
action: "stop"
function: "stop()"
Question:
state: "{} what should you do next?"
\end{lstlisting}

\section{Case Study}
\begin{figure*}[!t]
    \centering
    \includegraphics[width=1.0\linewidth]{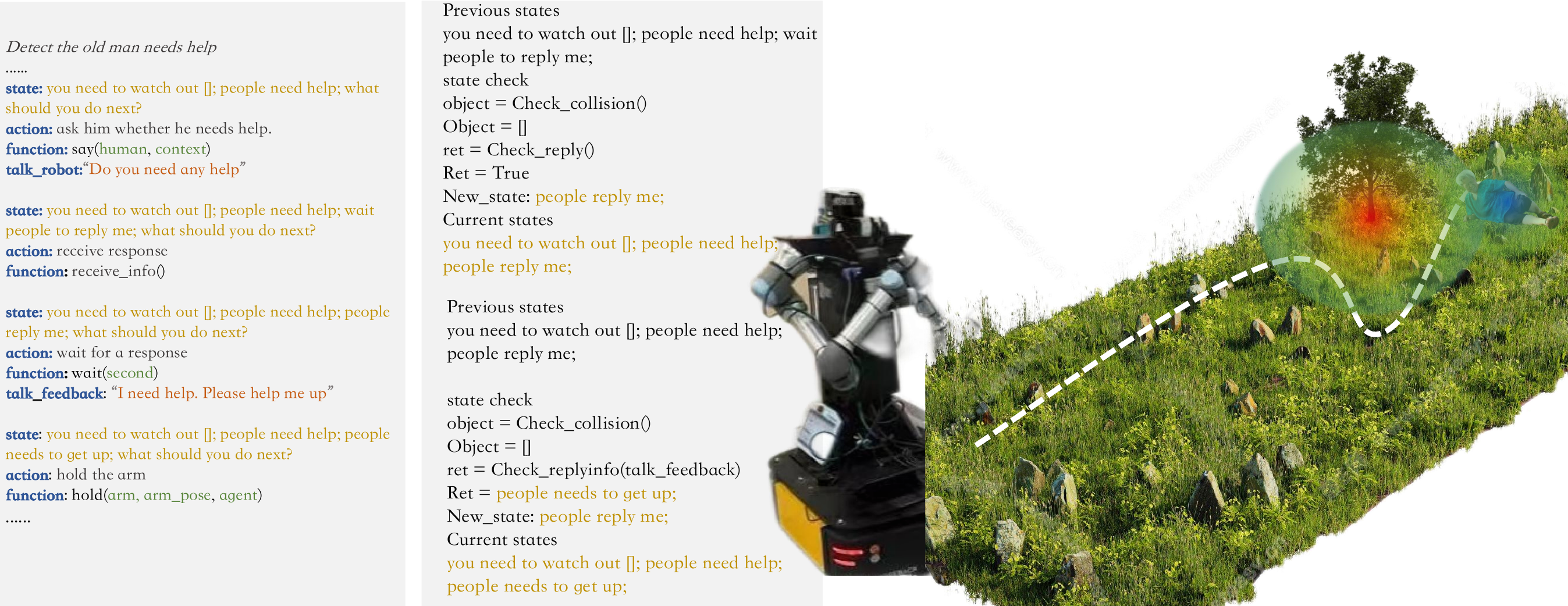}
   \caption{Case study of \model.}
    \label{fig:case_real_robot}
\end{figure*}

Fig.~\ref{fig:case_real_robot} shows a case study. \model effectively assesses the present condition and provides the appropriate handler function for the grip of the human arm. 

\begin{figure}[!t]
    \centering
    \includegraphics[width=1.0\linewidth]{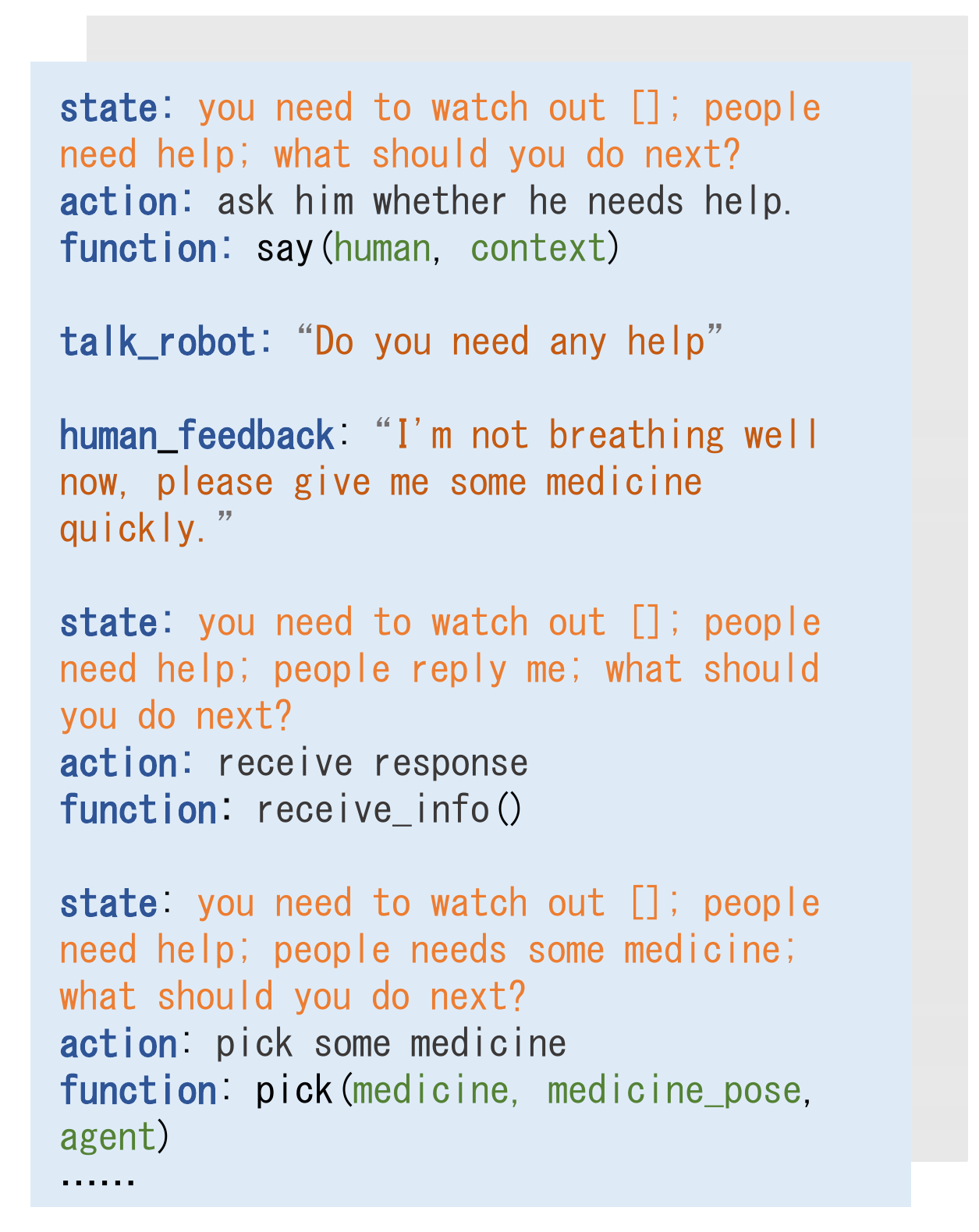}
    \caption{Task planning for ``\textit{Give patients  medicine}''.
    }
    \label{fig:plan_example}
\end{figure}

Fig.~\ref{fig:plan_example} shows a case study of planning.

\section{Ethics Statement}
This research will be conducted in accordance with the ethical principles of the American Psychological Association. We have received the RIB (Routing Information Base) assigned by Peking University. The research team will take all necessary steps to protect the safety and privacy of the participants. This includes obtaining informed consent from all participants, keeping participant data confidential, and monitoring participants throughout the research to ensure their safety. The research team will also consider the potential impact of the research on the participants and will design the research to minimize any potential harm.

\end{document}